\documentclass[10pt,twocolumn,letterpaper]{article}

\usepackage[pagenumbers]{cvpr} 

\usepackage[dvipsnames]{xcolor}

\usepackage{paralist}

\definecolor{cvprblue}{rgb}{0.21,0.49,0.74}
\usepackage[pagebackref,breaklinks,colorlinks,citecolor=cvprblue]{hyperref}
\usepackage{xcolor,colortbl}

\def\paperID{} 
\def\confName{}
\def\confYear{}

\title{\textit{SpatialBot}: Precise Spatial Understanding with Vision Language Models}

\author{Wenxiao Cai$^{1,2,3}$\thanks{This work was done when Wenxiao Cai was a visiting student in SJTU, and when he was an intern in BAAI.} , \hspace{0.5em}Iaroslav Ponomarenko$^{4}$, \hspace{0.5em}Jianhao Yuan$^{5}$, \hspace{0.5em}Xiaoqi Li$^{4}$,\\ Wankou Yang$^{6}$, 
\hspace{0.5em}Hao Dong$^{4}$, 
\hspace{0.5em}Bo Zhao$^{1,3}$\thanks{Corresponding author: bo.zhao@sjtu.edu.cn .} \\
\\
$^1$School of Artificial Intelligence, Shanghai Jiao Tong University  \hspace{0.7em} $^2$Stanford University \\ $^3$BAAI \hspace{0.7em} $^4$Peking University \hspace{0.7em} $^5$University of Oxford\hspace{0.7em} $^6$Southeast University \hspace{0.7em}  \\
}

\begin{document}
\maketitle
\begin{abstract}
Vision Language Models (VLMs) have achieved impressive performance in 2D image understanding; however, they still struggle with spatial understanding, which is fundamental to embodied AI.
In this paper, we propose \emph{SpatialBot}, a model designed to enhance spatial understanding by utilizing both RGB and depth images.
To train VLMs for depth perception, we introduce the \emph{SpatialQA} and \emph{SpatialQA-E} datasets, which include multi-level depth-related questions spanning various scenarios and embodiment tasks.
\emph{SpatialBench} is also developed to comprehensively evaluate VLMs' spatial understanding capabilities across different levels.
Extensive experiments on our spatial-understanding benchmark, general VLM benchmarks, and embodied AI tasks demonstrate the remarkable improvements offered by \emph{SpatialBot}. The model, code, and datasets are available at \url{https://github.com/BAAI-DCAI/SpatialBot}.
\end{abstract}    
\section{Introduction}
\label{sec:intro}

Recently, Vision Language Models (VLM)~\cite{gpt4v,gemini,llava,blip2,qwenvl,bunny} have demonstrated notable capabilities in general 2D visual understanding and reasoning, based on vision encoder-based perception and language model-based reasoning. However, it is still challenging for VLMs to comprehend spatial information from 2D images merely, which is the key to implementing various real-world tasks~\cite{sceneunderstand1,sceneunderstand2,3drecon1,3drecon2,compphoto1}, particularly those embodied AI related tasks such as manipulation~\cite{manip1, rtx, manip2, manip3} and navigation~\cite{navi1,navi2,navi3, nav4}. 

\begin{figure}[h]
	\begin{center}
    \includegraphics[width=1\linewidth]{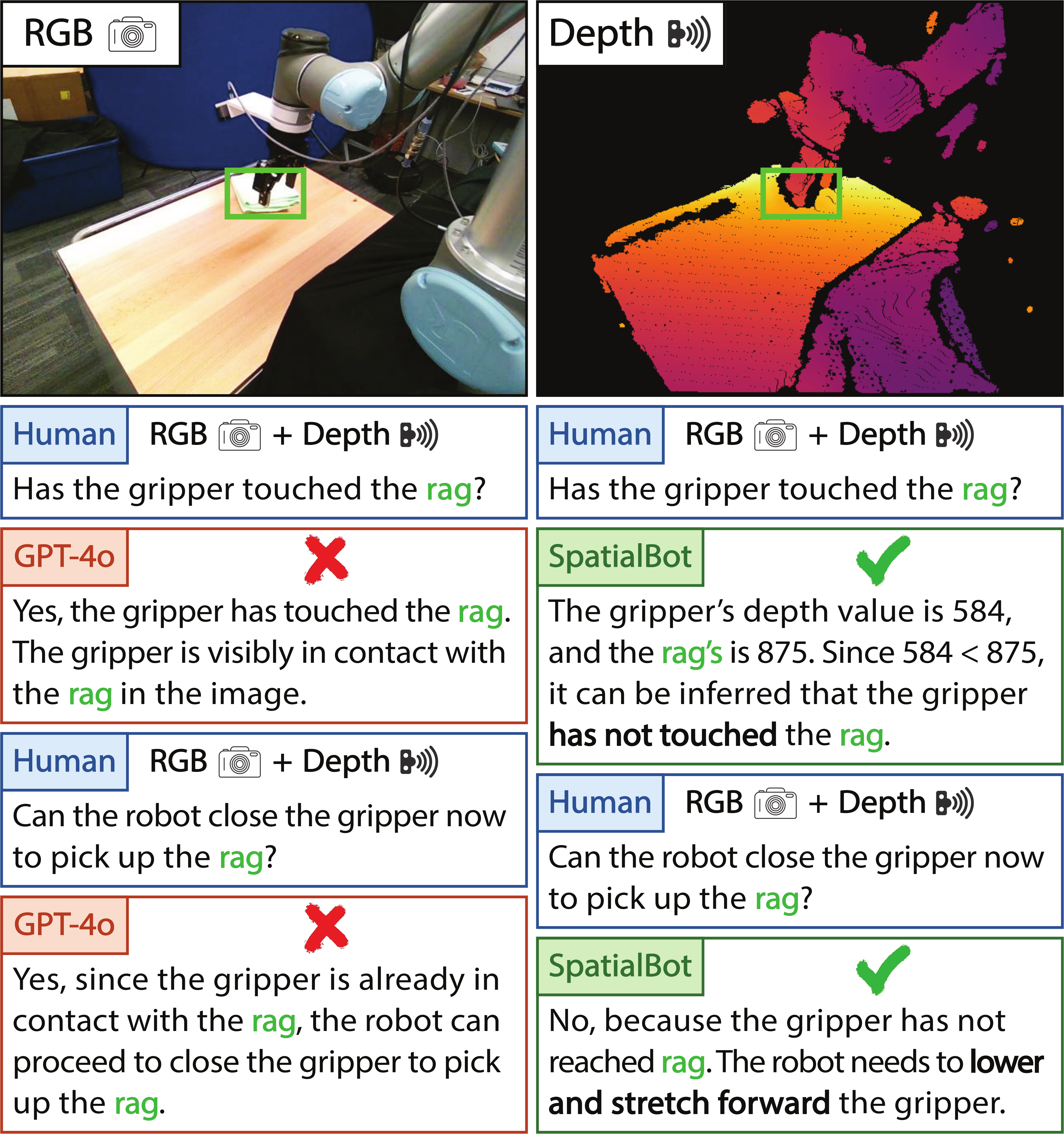}
	\end{center}
 	\caption{\textit{SpatialBot} has better spatial understanding ability than GPT-4o. From a single RGB image, even a human can not judge whether the gripper has touched the rag. GPT-4o fails even if looking at both RGB and depth images. \textit{SpatialBot} first obtains depth information of target objects from the depth map, and then judges that the gripper has not yet touched the rag.}
	\label{fig:depth_matters}
\end{figure}

The main challenges for VLMs to have spatial understanding ability are in the following aspects:
1) Popular VLMs have limited capacity to understand depth information as they are only trained on RGB images without seeing depth images. In addition, the training tasks need little depth information to solve. Consequently, directly inputting depth maps into VLMs results in poor performance.
2) A well-designed dataset for training VLMs to understand depth is absent. The popular VLM tuning datasets provide neither depth maps nor depth-related tasks.
3) The inconsistency of the scales between indoor and outdoor numerical depth is also an important problem preventing VLM from uniformly processing depth in various tasks. For example, tasks such as indoor navigation and manipulation require millimeter-level precision, whereas outdoor tasks do not necessitate such high precision but demand a broader depth range. 


To address these challenges, we propose \textit{SpatialBot}, which can precisely comprehend spatial information through depth images and perform robotic manipulation. We design a progressive training approach to first improve the general spatial understanding capacity of VLMs with the proposed \textit{SpatialQA} dataset, which contains general conversation tasks. We then leverage this spatial understanding capacity for embodied tasks using the collected robot manipulation task dataset, \textit{SpatialQA-E}.
We design various purpose-specific QA tasks that heavily rely on spatial understanding from low to high levels. These tasks include low-level depth estimation, middle-level object detection, referring QA, and depth comparison, high-level tasks that require depth reasoning, such as understanding spatial relationships in both general conversations and robot manipulation. 
To enable the model to accurately obtain depth information, we designed a depth API that allows the model to query the depth values of individual pixels or regions.

We validate the spatial comprehension capacity of VLMs with \textit{SpatialBench} which consists of manually annotated question-answer pairs on spatial understanding and reasoning. We also deploy \textit{SpatialBot} on robots to do manipulation tasks, for example, picking up the teacup in the middle and placing it on the closest board, as shown in Fig.~\ref{fig:teaser2}.
The experimental results verify that our SpatialBot can understand the depth in the three levels. Furthermore, it is also verified that the fine-tuning of VLMs in SpatialQA can improve their performance on general VLM benchmarks such as MME~\cite{mme}, MMBench~\cite{mmbench}, etc. Finally, robot manipulation abilities demonstrate the promising applications of \textit{SpatialBot}. 
In summary, the main contributions of our work are as follows:
\begin{compactitem}
\item We propose \textit{SpatialBot} that shows promising performance in general visual recognition, spatial understanding, and robot manipulation. 
\item We curate a large-scale RGB-D VQA dataset, \textit{SpatialQA}, for training \textit{SpatialBot}, and \textit{SpatialBench} for evaluating VLMs' spatial understanding performances. Three levels of tasks have been designed for a comprehensive analysis of depth.
\item We finetune and deploy \textit{SpatialBot} on embodiment tasks that involve spatial reasoning, and release the robot manipulation dataset focusing on spatial relationships, namely \textit{SpatialQA-E}.
\end{compactitem}

\begin{figure*}[t!]
	\begin{center}
    \includegraphics[width=1\linewidth]{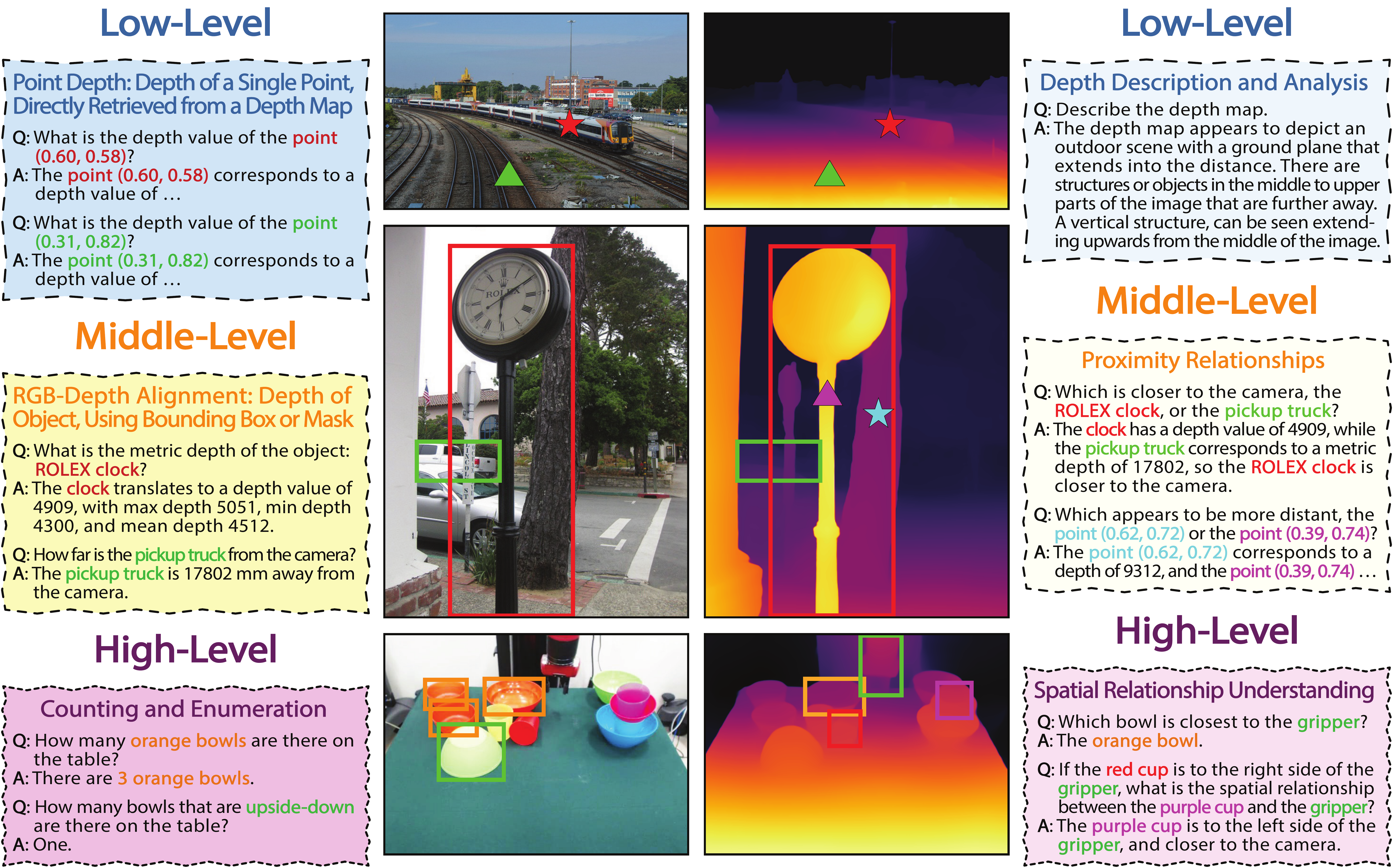}
	\end{center}
 	\caption{The proposed \textit{SpatialQA} dataset consists of basic, middle, and high-level VQAs in general VLM tasks, aiming to (a) help VLMs understand depth images, (b) let VLMs learn to align RGB and depth images, (c) enable VLMs to do high-level tasks better by understanding both RGB and depth images, as depth images provide clear boundary information and spatial relationships.}
	\label{fig:teaser}
\end{figure*}

\section{Related work}
\label{sec:relatedwork}

\subsection{VLM and RGB datasets}
In recent years, VLMs (or Multi-modal Large Language Models, MLLMs) have achieved significant advancements~\cite{efficientmllmsurvey}. LLaVA~\cite{llava} pioneered the visual instruction tuning, which is followed by subsequent works~\cite{qwenvl,yi-vl,bunny,mm1,minigemini} with more extensive datasets~\cite{svit} and different Large Language Models (LLM) backbones~\cite{llama,phi3,vicuna,qwen}. These VLMs primarily tackle tasks related to perception~\cite{mme}, reasoning~\cite{mmbench} and OCR~\cite{seedbench,mmmu}. Additionally, some works have introduced an encoder-decoder structure beyond VLMs to perform pixel-level grounding tasks~\cite{llava-grounding,ferret,lisa,osprey,nextchat,virl}. However, their performances in counting and enumeration~\cite{lamm,hallucinations} and spatial relationship understanding~\cite{vcoder} are mediocre. We posit that comprehending the entire space from a monocular RGB image is overwhelming for VLMs. Integrating depth information could effectively enhance the spatial understanding capabilities of VLMs.

\subsection{Spatial Understanding in General QA and Embodiment}
Spatial understanding requires VLMs to understand scenes beyond 2D RGB images. This is particularly crucial in precision tasks such as robotic grasping~\cite{graspnet}. Spatial understanding can be achieved through point clouds~\cite{graspnet,manifoundation} or depth maps~\cite{vcoder}. Some studies have attempted to perform depth estimation~\cite{proximityqa} and 3D detection~\cite{cubellm} directly from monocular RGB images, but the accuracy is limited regarding metric depth estimation.
SpatialVLM~\cite{spatialvlm} and SpatialRGPT~\cite{spatialrgbt} infer spatial relationships only from 2D images. However, in robotic tasks (see, e.g., Fig.~\ref{fig:depth_matters}), depth information from sensors is essential for spatial understanding.
Recently, Monocular Depth Estimation (MDE) has seen rapid advancements. Using large amounts of unsupervised data~\cite{depthanything,zoe} and synthetic data~\cite{diffusionmde}, MDE can accurately estimate the depth in various scenarios~\cite{midas}. Therefore, we improve the spatial understanding of VLMs by adding depth information to the RGB images they use, leveraging MDE.
Despite the strength of monocular depth estimation models, training large models to estimate depth directly is not always feasible. In embodied AI scenarios, precise depth information is required from hardware devices, which depth estimation models cannot achieve. Additionally, enabling VLMs to precisely understand space from a single RGB image has proven to be extremely difficult~\cite{proximityqa,cubellm}.
To extend spatial understanding abilities to embodiment, we propose \textit{ spatialQA-E}. To the best of our knowledge, it is the first manipulation dataset that focuses on spatial relationships.
\textit{SpatialBot} utilizes a similar model structure with state-of-the-art vision-language-action models like RT~\cite{rt1,rt2}, Octo~\cite{octo} and OpenVLA~\cite{openvla}, while acquires spatial knowledge necessary in manipulation tasks through training on \textit{SpatialQA-E}.
\begin{figure*}[h]
	\begin{center}
    \includegraphics[width=1\linewidth]{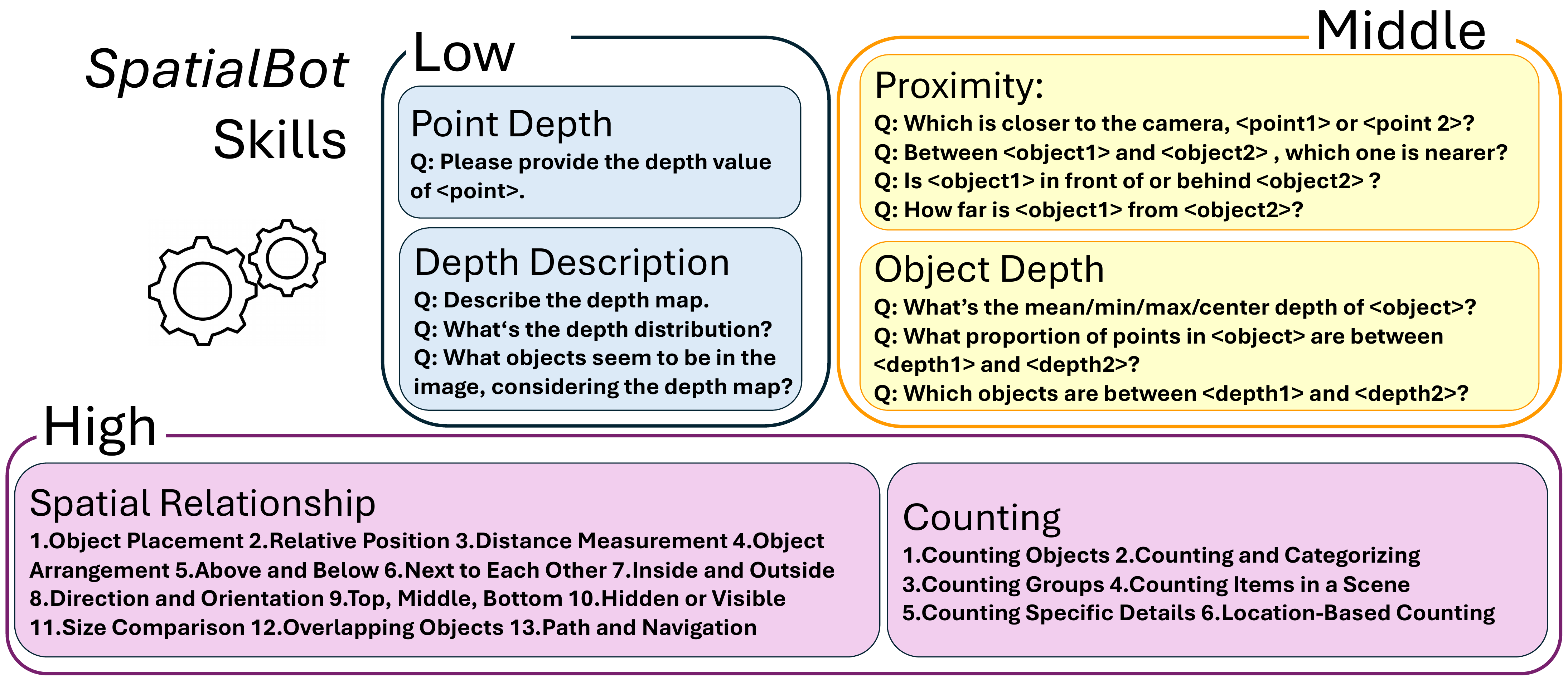}
	\end{center}
 	\caption{\textit{SpatialBot} masters three levels of skills: (a) understanding depth images (b) aligning these RGB and depth inputs, and performing proximity comparison, (c) applying RGB-D for spatial relationship understanding and counting.}
	\label{fig:skills}
\end{figure*}

\section{SpatialBot}
\label{sec:dataset}
We use depth information to guide VLMs in understanding space~\cite{depth3d1,depth3d2}, because compared to point clouds, depth information is easier to collect and process. Since the RGB-D cameras are cheap, most of robots carry such cameras to capture RGB and depth images instantly. In addition, due to remarkable capacities of Monocular Depth Estimation (MDE), one can adapt large scale RGB datasets to RGB-D dataset in a fast and affordable way.
Thus, we introduce depth images for spatial understanding and construct \textit{SpatialQA} dataset with RGB-D images and depth-related QA pairs. In this section, we elaborate on the pipeline of dataset construction: collecting RGB and depth images, estimating depth images from RGB images using MDE, unifying the format of depth images, generating basic VQAs for VLM training, and generating depth related VQAs. This pipeline can be easily scaled up to construct larger datasets from available RGB datasets.

\subsection{Depth Map Encoding}
Our depth encoding aims to preserve all depth information for VLMs to use. A challenge is the indoor and outdoor consistency. Indoor scenes like robot manipulation~\cite{rtx} and indoor navigation~\cite{navigation1,navigation2} may require millimeter-level precision, while outdoor scenes include a large range of depth values. Existing methods often adopt ordinal encoding~\cite{or,depthanything}, which, however, cannot be subjected to basic mathematical operations. 
To address the issue, we use uint24 or three-channel uint8 to store depth values, measured in millimeters from $1mm$ to $131.071m$. We directly save the raw depth values and leave subsequent computations to the powerful fitting capabilities of VLMs. For single-channel uint24, we use millimeter as unit directly. This way, VLMs can directly query the required values from the depth map.
For three-channel uint8 images, we distribute the values across a broader range: the units for the three channels are $2^0$, $2^5$, and $2^{10}$ millimeters, respectively. Each channel has $2^5$, $2^5$, and $2^7$ possible values. For an image of size $(H,W)$, to store depth value $d_{H,W}$ (in millimeters) in three-channel uint8 image $I_{H,W}^{3}$, we encode the image $I$ following:
\begin{align}
    I_{h,w}^{0} &= ( d_{h,w} // 2^{10} ) *2^1, \\
    I_{h,w}^{1} &= ( d_{h,w} // 2^5 ) *2^3, \\
    I_{h,w}^{2} &= ( d_{h,w} \% 2^5 ) *2^3.
\end{align}
The choice of $2^{10}$ mm as a unit for the first channel is influenced by the depth range in many desktop grasping tasks in robotics~\cite{ur5,berkerlybridge,ucsdpickplace,stfrobocook,uiucd3field,droid,vima}, which typically have a maximum depth of around $1m$. A larger unit would result in the first channel being predominantly zero in most scenarios.
Similarly, we use multipliers of 2 and 8 to ensure the  better distinction between three-channel depth map. We believe VLMs can easily learn the relationship between our encoding method and the actual depth values, and our experiments have validated this.

If the raw data includes depth estimated by sensors, we use the raw depth values. Otherwise, for MDE, we use the ZoeDepth~\cite{zoe} model for estimation, as it considers both indoor and outdoor scenarios and can accurately estimate metric depth in these situations.
Note that we do not use the relative depth models, such as MiDaS~\cite{midas}. It is incorrect to directly take the inverse of the relative depth $d_r$ as the actual depth $d$. Suppose that the maximum and minimum depth in an image is $d_{max}$ and $d_{min}$ ,the conversion should follow:
\begin{align}
    A &= \frac{1}{d_{min}} - \frac{1}{d_{max}}, \\
    B &= \frac{1}{d_{max}}, \\
    d &= \frac{1}{(A * d_r + B)}.
\end{align}

It is incorrect to ask for the depth value but uses $\frac{1}{d_r}$ as the label in QAs~\cite{proximityqa}. While $\frac{1}{d_r}$ can reflect the relative size of depth, it does not maintain proportional relationships (e.g. $\frac{1}{d_r} = 0.4$ is not twice the depth of $0.2$). Only when the max depth is infinite can $\frac{1}{d_r}$ be considered as true depth multiplies by a scale factor.

\subsection{Depth Description of an Object}
\textit{SpatialQA} is a VQA dataset, and our model is a standard VLM (Fig.~\ref{fig:model}): it takes images and text as input and outputs text. 
To maintain generality, we do not use a separate image encoder, so \textit{SpatialBot} cannot output pixel-level information. Intuitively, the center point of objects can simply represent their depth. However, for example, in the case of a cup, there is a significant difference between the depth of the inner and outer surfaces, so a single value cannot accurately represent the depth.
Therefore, we use four depth values—max, min, mean, and center—to describe the object's depth, if its mask is available. Considering that the mask and depth map cannot be perfectly precise, we use the 95th and 5th percentile values as the max and min depth values. 
Bounding boxes in Visual Genome (VG)~\cite{vg} are very inaccurate, and our experiments find that prompting SAM~\cite{sam} with these bounding boxes will yield undesirable masks. 
In this case, to prevent incorrect depth from misleading the model, we fall back to using only the depth of the center point of the bounding box to describe depth.

\subsection{Image Sources}
\begin{figure*}[h!]
	\begin{center}
    \includegraphics[width=1\linewidth]{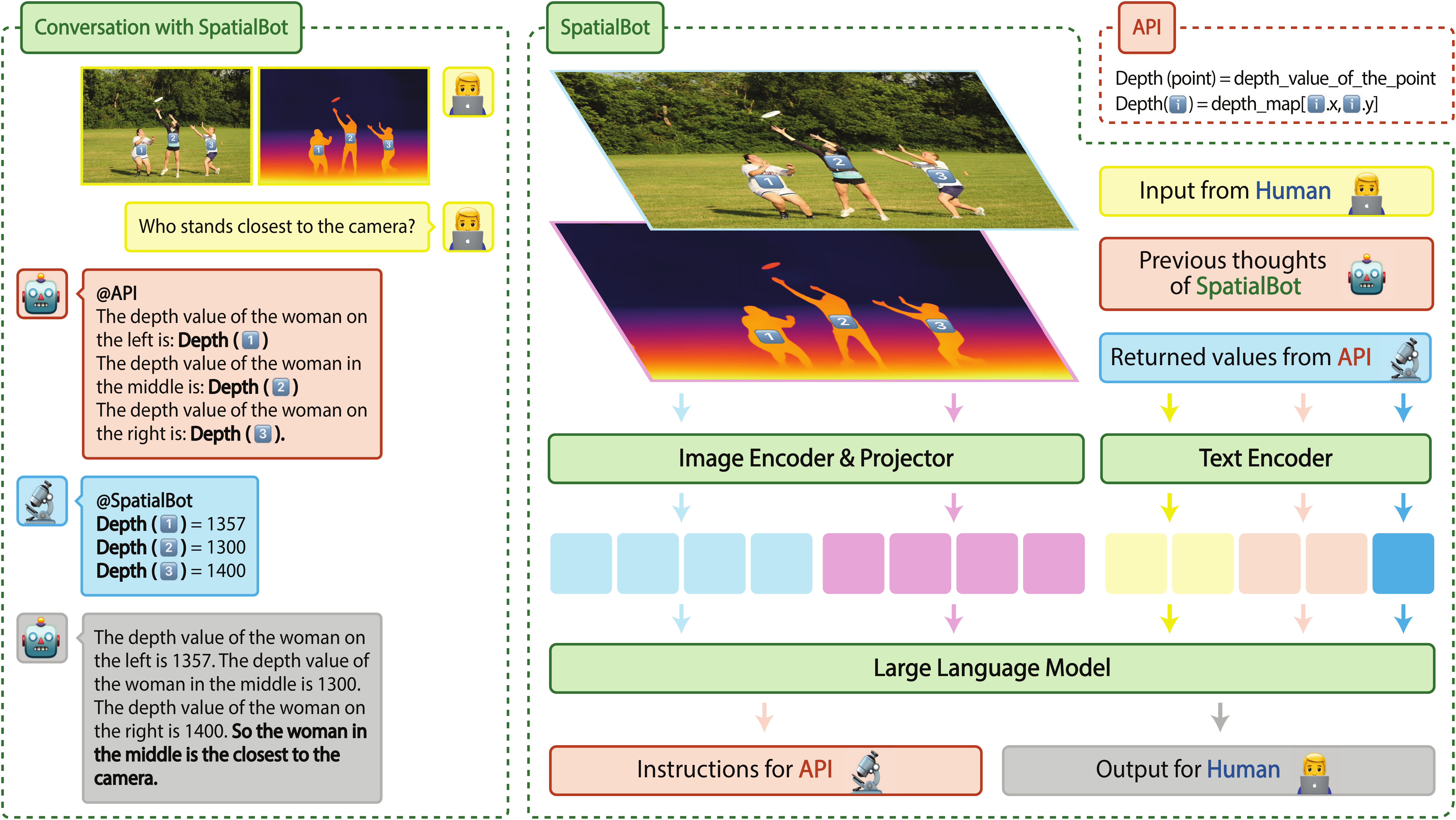}
	\end{center}
 	\caption{The architecture of \textit{SpatialBot}. It takes a pair of RGB and depth images as input, where depth images are optional. \textit{SpatialBot} can choose to call Depth API if it need accurate depth information.}
	\label{fig:model}
\end{figure*}

A RGB-D VLM dataset should include detailed QAs that help VLMs to understand the image, which may include reasoning, conversation, description and referring~\cite{svit}. Specific object descriptions are required, e.g. in Fig.~\ref{fig:model}, woman is not a good description, but the woman in the middle or the woman standing tallest are good descriptions. Existing captioning, grounding and segmentation models~\cite{grounding-dino,yolo-world,grounded-sam,seem} can not generate detailed and specific descriptions. To this end, we base \textit{SpatialQA} mainly on VLM data where detailed QAs are included~\cite{bunny}. 

In \textit{SpatialQA}, we primarily include three data sources: COCO~\cite{coco}, VG~\cite{vg}, and Open X-Embodiment (RTX)~\cite{rtx}. 
Therefore, we base our dataset on Bunny\_695k~\cite{bunny}, which includes COCO and VG. 
Bunny\_695k contains image QA covering reasoning, detailed descriptions, grounding, etc. On this basis, we added depth-related QA pairs.
We use bounding boxes in Bunny\_695k and prompt SAM~\cite{sam} with bounding boxes and center points to get masks.
We ensure that SAM masks do not exceed the bounding box limits, then select the mask with the highest confidence.
RT-X integrates many robotics datasets. For datasets containing sensor depth data, we directly use the raw depth. For other datasets, we use model-estimated metric depth. We select 7.5k of these and manually annotated the bounding boxes, querying the depth information of the objects. For the remaining images, we only ask about the depth of certain pixels in the image.
Also, we use GPT-4o to generate conversations based on RTX-7.5k, where we prompt GPT-4o to focus on: what robot are doing, how should the robot finish robot task, object count, object position, positional relationships and object appearance.
In future versions of \textit{SpatialQA}, we will include more images from a vast range of sources.

\begin{figure*}[h]
\vspace{-2em}
	\begin{center}
    \includegraphics[width=0.95\linewidth]{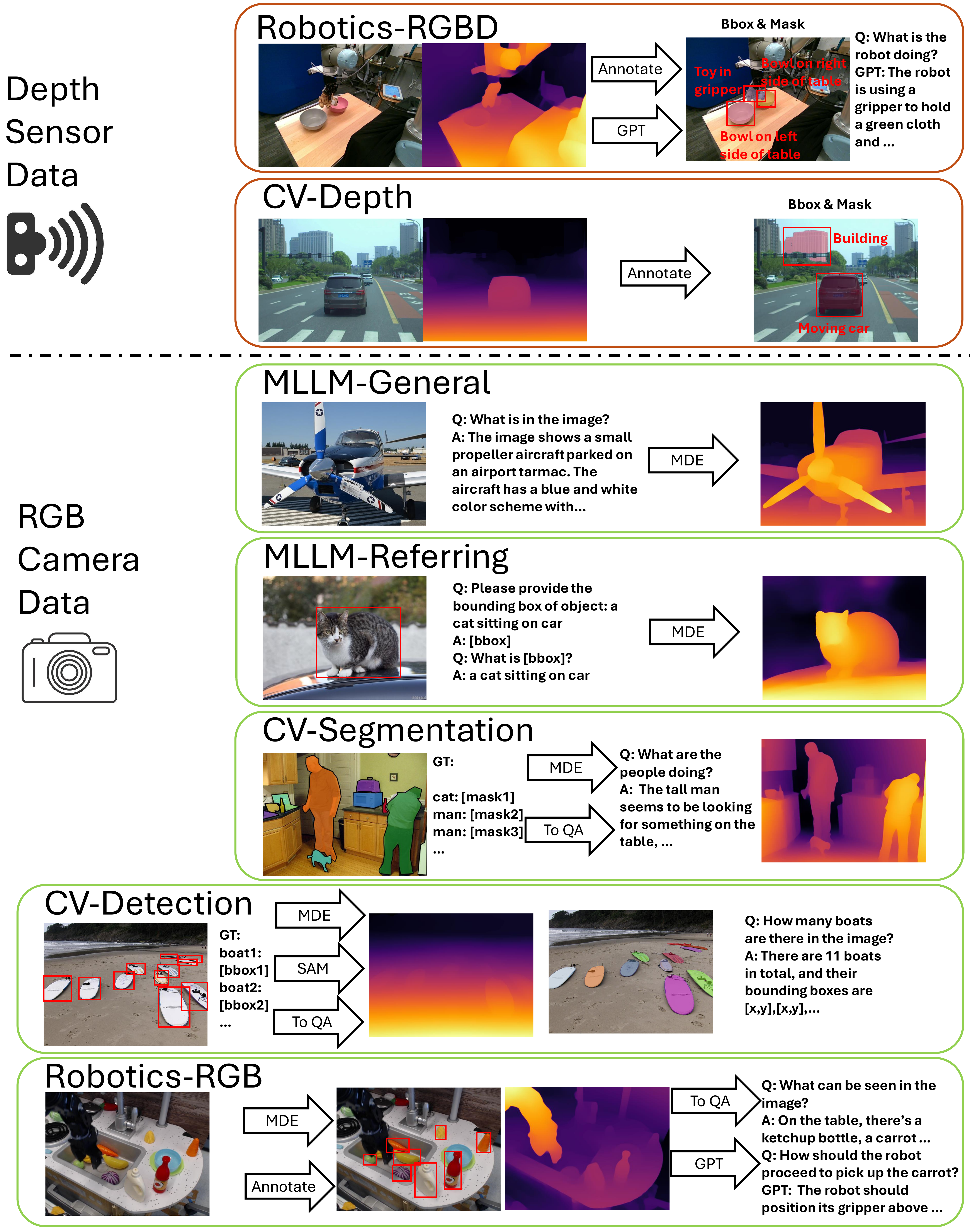}
	\end{center}
 	\caption{Image sources and RGB to RGB-Depth dataset conversion pipeline. RGB and depth information, captions or conversations about the images, bounding boxes or masks, and corresponding detailed descriptions of objects are required to make \textit{SpatialQA} dataset.}
	\label{fig:img_source}
\end{figure*}

\subsection{\textbf{\textit{SpatialQA}} Pipeline}
To help VLMs understand depth inputs, and use depth information to do high-level tasks like spatial relationship understanding, counting and enumeration, we design a three-step QA pipeline. We aim to make this pipeline effective and easy-to-follow: 
(a) This pipeline progressively let VLM learn to understand depth, align depth and RGB, and use depth for complex reasoning in high-level tasks.
(b) existing RGB datasets can be easily converted to RGB-Depth datasets with our pipeline. \textit{SpatialQA} pipeline is shown in Fig.~\ref{fig:img_source}, and the skills to learn are shown in Fig.~\ref{fig:skills}.

\textbf{Low level}. To enable VLMs to understand depth images and learn to query information from them, we ask depth value of points. VLMs should learn to take the depth value directly from depth inputs, and relate point coordinates with pixels in image. In the meanwhile, since the visual encoder does not see depth images in pre-training, we also expect the encoder and projector to learn to encode depth images together with RGB images.
We also let \textit{SpatialBot} describe the depth map and infer what may be in the images, giving only a depth map.

\textbf{Middle level}. As VLMs have learnt to encode and query information from depth images, they should now learn to use depth information. Also, since image and depth inputs are given to VLMs, they should also know the relationships between them. First, we ask about proximity relationships, namely which point is closer or further away. Second, we let VLMs learn to describe the depth of objects or regions, by using center point depth, minimum, maximum and mean depth. VLMs should learn to locate an object in the RGB image and then find depth information from depth input.
Third, we ask about proximity relationships between objects. 

\textbf{High level}.
Since VLMs can now understand depth input, align depth with RGB and have some knowledge about proximity relationships in the spatial world, we design tasks to help VLMs apply depth at a higher level. 
When the model sees the depth map, the boundaries of objects and their surroundings become clearer, so we believe that the depth map aids in grounding and counting tasks. Additionally, in \textit{SpatialQA}, the model gains a clear understanding of the space, which helps the model determine spatial and positional relationships.

\subsection{\textbf{\textit{SpatialQA-E}}}
\begin{figure}[h]
	\begin{center}
    \includegraphics[width=1\linewidth]{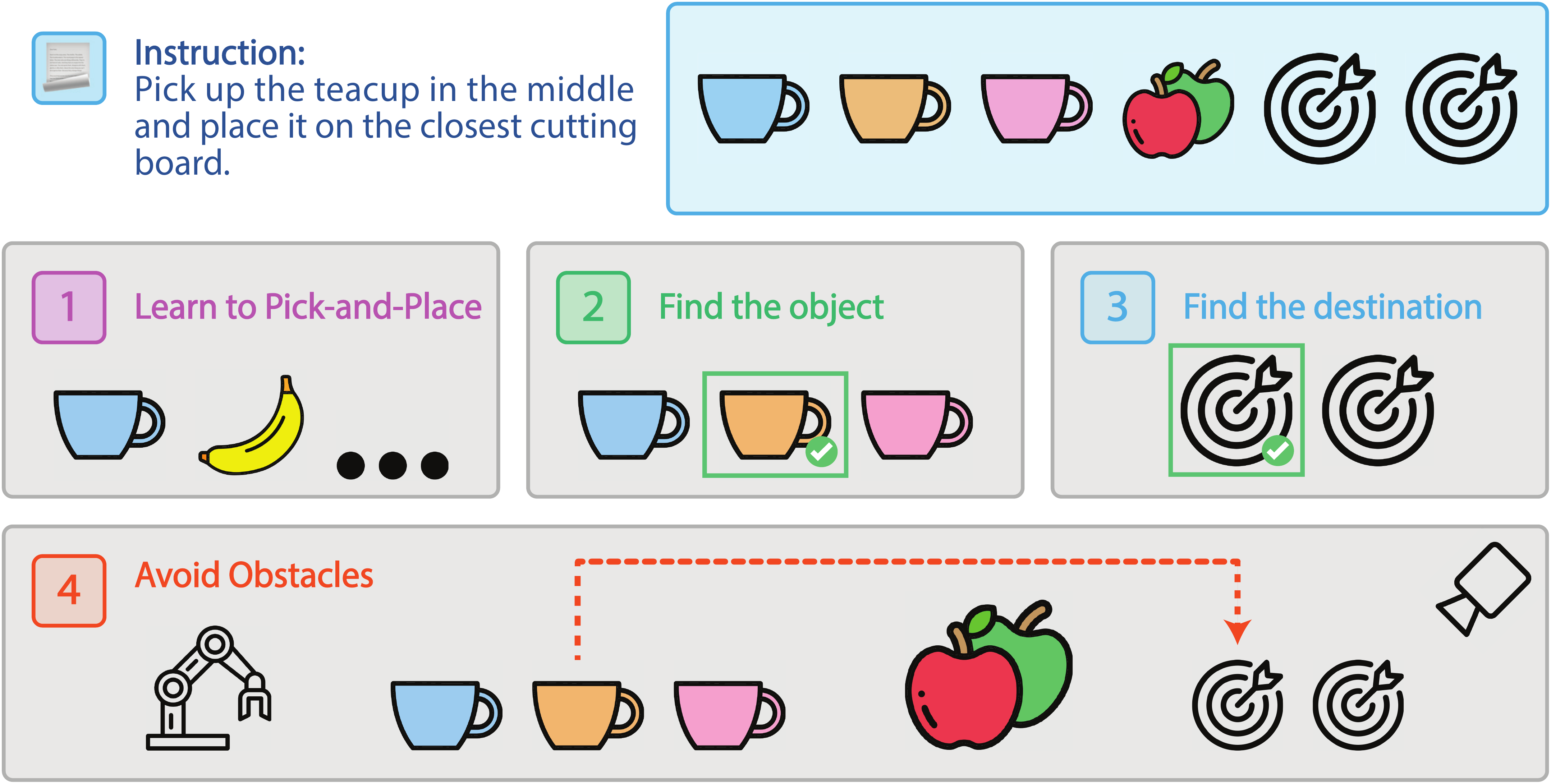}
	\end{center}
 	\caption{\textit{SpatialQA-E} involves spatial relationships in robot manipulation.}
	\label{fig:teaser2}
\end{figure}

\begin{figure}[h]
	\begin{center}
    \includegraphics[width=1\linewidth]{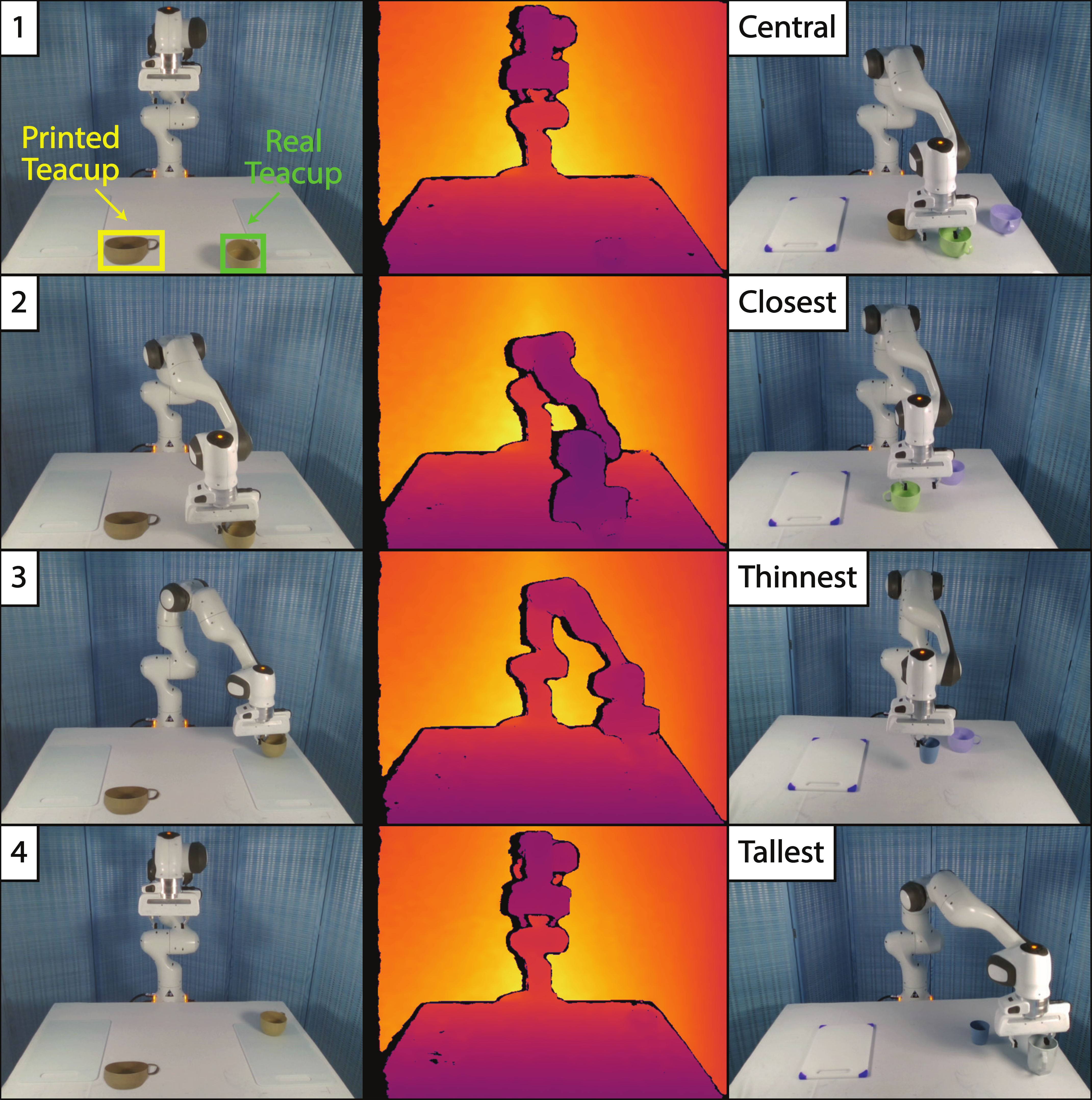}
	\end{center}
 	\caption{\textit{SpatialQA-E} demonstration. Left: 4 steps in picking up the real teacup and putting it on the right cutting board relative to the camera. We print the teacup as a distraction. It's easier to tell between the real and printed teacup from the depth map. Right: 4 sample settings in \textit{SpatialQA-E}, where we specify spatial relationships.}
	\label{fig:spatialqa-e-demo}
\end{figure}

We propose \textit{SpatialQA-E} to extend spatial understanding and reasoning abilities to embodiment tasks. We use the 7-axis Franka Research 3 Robotic Arm to grasp objects on the table, avoid obstacles while moving, and place them on a cutting board on the table. We include spatial relationships in language instructions, so the model should learn spatial reasoning in manipulation. 
\textit{SpatialQA-E} contains 2000 episodes in total. 
The dataset is composed of 4 steps, shown in Fig.~\ref{fig:teaser2} and Fig.~\ref{fig:spatialqa-e-demo}:
\begin{compactitem}
    \item Learn to pick and place teacups, balls, bananas, etc.
    \item Find specific object and destination. The dataset includes spatial relationships in positive, comparative (-er), and superlative (-est) degrees from the perspective of the robot or the human (camera): 
    \begin{compactitem}
    \item Positional: left/ right/ middle/ up/ down on/ in/ inside/ outside
    \item Size: tall/ short/ large/ small/ wide/ thin/ big/ small
    \item Illusion: we take photos of objects, print them out, and put the printed object on the table. It looks real, and the model needs to tell between printed and real objects through visual clues, e.g., depth information (printed objects are flat) and shadows.
    \end{compactitem}
    \item In moving objects, the robot needs to avoid obstacles.
\end{compactitem}

\subsection{SpatialBot in Embodiment Tasks}
\textit{SpatialBot} is finetuned on \textit{SpatialQA-E} to work on embodiment tasks. In short, it is a Vision-Language-Action (VLA) model that supports multi-frame RGB or RGB-D inputs.
Robot manipulation tasks are specified as: in current time stamp $t$, given history and current image observations ${x_j}_{j=0}^{t}$ (RGB or RGB-D), models should learn policy $\pi(·|i,{x_j}_{j=0}^{t})$. Action $a_t$ is sampled from $\pi$ and applied to robots. For a robot of two-finger end effector, action space can be represented as 7 DoF vector: $(\Delta X, \Delta Y, \Delta Z, \Delta R, \Delta P, \Delta Yaw, C)$, indicating delta change in poses $XYZ$ and rotation $RPYa$ (roll, pitch, yaw), gripper closure $C$. The delta change in position and rotation of action space is encoded into 101 possible values, from 0, 0.01 to 1. The model output texts of 7 DoF actions directly. A sample conversation: 'User: What should the robot do to pick up the biggest teacup and move it to the left cutting board? Answer with robot parameters. - SpatialBot: The robot should $<$0.17, 0.51, 0.44, 0.62, 0.83, 0.07, 1$>$'. Then we decode the output to robot control signals to control the robot movement of each frame.
If the model directly answers robotic parameters during the finetuning stage, we find that it can only respond to robot-specific questions. To enable multi-task training, we incorporate some natural language elements into the robot's responses, such as 'The robot should'. Then, we train the model on robotic data and general QA data, such as SpatialQA-E and SpatialQA.
We have the model predict special tokens during robotic tasks to maintain the model's numerical reasoning abilities in general conversations. 
We predict each frame's delta pose instead of the target pose. This choice allows for more precise control of the robot by dividing each dimension of the action space into 100 bins. 
Additionally, we found that delta pose is harder to learn than target pose, as reflected in the slower decrease in loss. For some data, the delta pose loss doesn't decrease at all, and the model completely collapse and output the same value regardless of the input. A quick solution to this issue is to exponentially increase the amount of training data.

\subsection{\textbf{\textit{SpatialBench}}}
To evaluate VLM's performance on high-level tasks, we annotate the \textit{SpatialBench}. On 120 images, we ask following questions:
\begin{compactitem}
    \item Has [Object 1] touched/reached [Object 2]?
    \item What is the spatial relationship between [Object]s?
    \item Counting and enumeration.
    \item Size comparison between objects.
\end{compactitem}
All question are in Yes/No or multiple choice formats. 
Additionally, \textit{SpatialBench} includes depth and proximity questions on images from MME~\cite{mme} dataset, and our manually annotated 80 images (3 bounding box per image).

\subsection{\textbf{\textit{SpatialBot}} Depth API Architecture}
\textit{SpatialBot} uses a VLM structure: Images are processed through an image encoder and a multi-modal projector, converted into tokens, and then sent along with text tokens into an LLM, which ultimately outputs responses. 
To enable the model to accurately obtain depth information, we designed \textbf{Depth API}.
When the \textit{SpatialBot}'s output contains text with a format of Depth(point), the API will query the depth value of that point in the corresponding depth map and then input this depth value back into \textit{SpatialBot}. Combining the user's question with the API's return value, \textit{SpatialBot} will provide the final answer.
The model can call the API to get the precise depth value of a specific point. For example, when \textit{SpatialBot} wants to know the depth information of an object, it first determines the bounding box of the object and then calls the Depth API using the center point of the bounding box. If the model wants to obtain the depth range of this object, it first observes which points in the image correspond to the maximum and minimum depth values and calls the Depth API using the coordinates of these points. However, to enhance the model's understanding of the depth map itself, during training, we only allow \textit{SpatialBot} to call the API on a subset of the data. For the remaining data, the model must directly use the depth map to answer the depth of the object.
\section{Experiments}
\label{sec:experiments}

\begin{table*}[h!]
    \centering
    \caption{Results on \textit{SpatialBench}. The best results of models with the same base LLMs are marked with \textbf{bold} text. LLM-RGB and LLM-RGBD are trained on RGB images only and tested with RGB and RGBD inputs, respectively. \textit{SpatialBot} with RGB input in depth estimation is the same as the MDE task.}
        \begin{tabular}{l | cccccccc}
            \toprule
            Model  & Depth $\uparrow$&  Position $\uparrow$& Existence $\uparrow$ & Counting $\uparrow$& Reaching $\uparrow$& Size $\uparrow$\\
            \midrule
            
            GPT-4o-RGB  & -  & 70.6 & 85.0 & 84.5  & 51.7 & 43.3 \\
            GPT-4o-RGBD  & - & 61.8 & 90.0 & 85.2 & 51.7 & 40.0 \\
            \midrule
            
            Bunny-Phi2-3B-RGB & 70.6 & 50.0 & 75.0 & 89.4 & 51.7 & 26.7 \\
            \rowcolor{lightgray} \textit{SpatialBot}-Phi2-3B-RGB & 84.1 & \textbf{64.7} & \textbf{80.0} & 88.0 & \textbf{61.7} & \textbf{28.3} \\
            Bunny-Phi2-3B-RGBD & 85.8  & 50.0  & 75.0  & 90.4 & 43.3 & \textbf{28.3} \\
            \rowcolor{lightgray} \textit{SpatialBot}-Phi2-3B-RGBD & \textbf{\textgreater99} & 61.8  & \textbf{80.0} & \textbf{91.7} & 55.0 & 26.7 \\

            \midrule

            Bunny-Phi3-4B-RGB & 32.3 & 58.8 & \textbf{75.0} & 91.0  & 31.7 & 16.7 \\
            \rowcolor{lightgray} \textit{SpatialBot}-Phi3-4B-RGB & 83.2 & 64.7 & \textbf{75.0}& 91.0 & \textbf{40} & \textbf{23.3} \\
            Bunny-Phi3-4B-RGBD & 63.3 & 52.9 & 60.0 & 85.4  & 31.7  & 18.3 \\
            \rowcolor{lightgray} \textit{SpatialBot}-Phi3-4B-RGBD & \textbf{\textgreater99} & \textbf{67.7} & 70.0 & \textbf{91.7} & 35.0 & 21.7 \\

            \midrule
            
            Bunny-QWen-1.5-4B-RGB & 42.2 & 50.0 & \textbf{75.0} & \textbf{91.6} & 26.7 & 15.0 \\
            \rowcolor{lightgray} \textit{SpatialBot}-QWen1.5-4B-RGB & 89.9 & \textbf{52.9}& \textbf{75.0} & 88.6 & \textbf{46.8} & 18.3 \\
            Bunny-QWen-1.5-4B-RGBD & 74.6 & 44.1  & 70.0  &  90.7 &  25.0 &  15.0 \\
            \rowcolor{lightgray} \textit{SpatialBot}-QWen1.5-4B-RGBD & \textbf{\textgreater99} & \textbf{52.9} & 60.0 & 90.5 & 41.7 & \textbf{26.7} \\

            \midrule
            
             Bunny-Llama3-8B-RGB & 58.1 & 50.0 & 75.0 & 91.7  & 38.3 & 23.3 \\
             \rowcolor{lightgray} \textit{SpatialBot}-Llama3-8B-RGB & 85.6 & \textbf{55.9} & \textbf{80.0} & \textbf{91.2} & 40.0 & 20.0 \\
             Bunny-Llama3-8B-RGBD & 64.0 & 50.0  & 75.0 & 90.4  &  38.3 & \textbf{25.0} \\
             \rowcolor{lightgray} \textit{SpatialBot}-Llama3-8B-RGBD & \textbf{\textgreater99} & 53.0 & 75.0 & 90.4 & \textbf{45.0} & 20.0 \\
    
            \bottomrule
        \end{tabular}
            
    \label{tab:spatialqa_benchmark}
\end{table*}

\begin{table*}[h!]
    \centering
    \caption{Results on general VLM Benchmarks. For the same base LLM models, better results are marked with \textbf{bold} text. RGB-D inputs are only used in MME. We report the results of Bunny trained with RGB and tested with RGB/RGB-D in it, split with slash. \textit{SpatialBot} is trained on RGBD and tested on RGB/RGB-D on MME.}
        \begin{tabular}{l | ccccccccc}
            \toprule
            Model  & MME$^\text{P}$ $\uparrow$ & MME$^\text{C}$ $\uparrow$ & MMB$^\text{T}$ $\uparrow$ & MMB$^\text{D}$ $\uparrow$ & SEED-I $\uparrow$ & VQA$^\text{v2}$ $\uparrow$ & GQA $\uparrow$ & POPE $\uparrow$ \\
            \midrule
            
            Bunny-Phi2-3B  & 1472/1474 & 286/285 & 67.90 & \textbf{68.90} & 69.91 & 78.98 & 61.52 & 86.21 \\
            
            \rowcolor{lightgray} \textit{SpatialBot}-Phi2-3B  & 1483/\textbf{1487} & 310/\textbf{312} & \textbf{70.12} & 68.56 & \textbf{70.85} & \textbf{79.80} & \textbf{62.28}  & \textbf{87.04} \\

            \midrule
            Bunny-Phi3-4B  & 1417/1364 & 308/319 & 70.15 & 70.74 & 71.04 & \textbf{80.57} & 61.18  & 84.60 \\
            
            \rowcolor{lightgray} \textit{SpatialBot}-Phi3-4B  & \textbf{1431}/1433 & \textbf{337}/329 & \textbf{73.49} & \textbf{73.11}  & \textbf{71.64} & 80.01 & \textbf{62.16} & \textbf{85.47} \\

            \midrule
            Bunny-QWen1.5-4B  & 1340/1364 & 251/254 & 69.56 & 68.56 & 70.05 & \textbf{80.63} & 61.55 & 85.10 \\
            \rowcolor{lightgray} \textit{SpatialBot}-QWen1.5-4B  & 1378/\textbf{1406} & 266/\textbf{285} & \textbf{70.91} & \textbf{69.67}  & \textbf{70.36} & 79.69 & \textbf{62.77}  & \textbf{86.09} \\

            \midrule
            Bunny-Llama3-8B  & 1574/1542 & 342/318 & 73.67 & 74.15 & 72.32 & 80.50 & 62.18  & 85.22 \\
            \rowcolor{lightgray} \textit{SpatialBot}-LLama3-8B  & \textbf{1577}/1576 & \textbf{352}/333 & \textbf{75.78} & \textbf{74.83}  & \textbf{72.40} & \textbf{80.94} &  \textbf{62.90} & \textbf{85.33}  \\

            \bottomrule
        \end{tabular}
    
    \label{tab:main}
 \vspace{-2em}
\end{table*}

We start with validating \textit{SpatialBot} has the ability to understand depth, extract information from depth maps, and perform high-level tasks.
Then, we observe performance improvements in general VQA tasks, such as MME~\cite{mme} and GQA~\cite{gqa}, by introducing depth maps. This indicates that training on \textit{SpatialQA} can help VLMs perform better on general tasks.
Finally, experiments on RT-X~\cite{rtx} show that \textit{SpatialBot} benefits from understanding depth in robot manipulation tasks.

\textbf{Implementation}. We design \textit{SpatialBot} based on Bunny~\cite{bunny}, a family of VLMs. Phi-2-3B, Phi-3-4B~\cite{phi3}, QWen-1.5-4B~\cite{qwen} and Llama-3-8B~\cite{llama} are used as the base LLM. \textit{SpatialBot} model architecture is shown in Fig.~\ref{fig:model}.
The image encoder is SigLIP~\cite{siglip} with 384x384 image resolution.
QWen-1.5-0.5B and CLIP~\cite{clip} with 336x336 image resolution are adopted in robot manipulation tasks.
We pretrain models on two million image-text pairs from LAION-2B~\cite{laion} (Bunny-pretrain-LAION-2M~\cite{bunny}) and finetune them on Bunny\_695k~\cite{bunny}. 
The learning rate is kept $2e-4$, and learning rate for multi-modal projector is $2e-5$, except for Llama-3-8B, where we halve both learning rates.
For manipulation tasks, we also halve the learning rate. The multi-modal projector is trainable in both pretrain and finetune stage, and we add a LoRA~\cite{lora} module in finetuning.
We use 8 A100 for training. On \textit{SpatialQA}, it takes about 15 hours for Phi-2~\cite{phi3}.

\subsection{Spatial Understanding}
\begin{figure*}[h!]
	\begin{center}
    \includegraphics[width=0.9\linewidth]{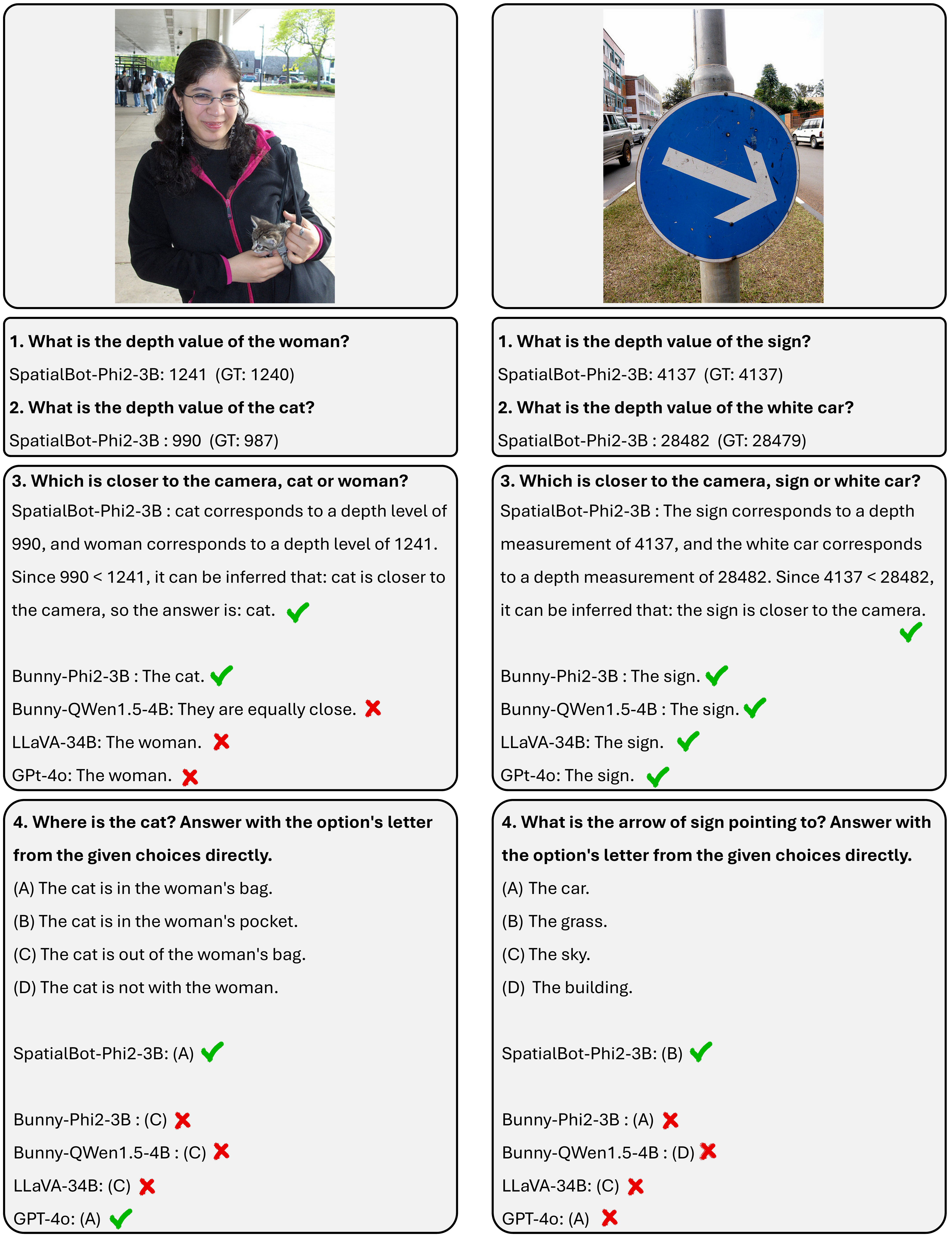}
	\end{center}
 	\caption{Sample conversations of \textit{SpatialBot} and baseline models. It first asks about depth value of objects, then lets models compare depth between objects. Finally, spatial relationship questions are asked.}
	\label{fig:conv1}
\end{figure*}

We first validate that \textit{SpatialBot} can get accurate metric depth value from depth images or Depth API, and decide proximity relationships, which are low-level and middle level tasks in \textit{SpatialQA}. 
We use bounding box and metric depth from \textit{SpatialBench}. We then ask about depth of random points and objects in them. 
We tell VLMs the names w/ and w/o bounding boxes of target object.
For ground truth depth value $d_{gt}$, estimated depth value $d_{est}$ from VLMs, we estimate depth accuracy by $\frac{d_{gt} - d_{est}}{d_{gt}} * 100\%$. 
Results by answering with Depth API are shown in Depth and Proximity column in Table.~\ref{tab:spatialqa_benchmark}.
Also, we ask the proximity relationships. Sample conversations on depth map understanding and high-level tasks are shown in Fig.~\ref{fig:conv1}.

\subsection{\textbf{\textit{SpatialBench}}}
We compare model performance on our \textit{SpatialBench}, which composes on positional relationship, object existence, reaching and size comparison tasks. GPT-4o is compared with models trained on \textit{SpatialQA}. 3B, 4B and 8B models trained on \textit{SpatialQA} reaches comparable results with GPT-4o. Results are reported in Table~\ref{tab:spatialqa_benchmark}.

\subsection{General VLM Benchmarks}
We report results on general benchmarks: MME perception~\cite{mme} (MME$^\text{P}$), MME cognition (MME$^\text{C}$), MMBench~\cite{mmbench} test and dev set (MMB$^\text{T}$ and MMB$^\text{D}$), SEED Bench Image~\cite{seedbench} (SEED(-I)), VQA~\cite{vqav2} test-dev split (VQA$^\text{v2}$), GQA~\cite{gqa}, and POPE ~\cite{pope} (the averaged F1-score of three categories on the validation set of COCO). In most of these benchmarks, RGB information along is enough. We only use RGB-Depth input on MME$^\text{P}$ and 
GQA since they contain counting, existence and position questions, where we expect depth information can benefit such cases. 

\subsection{SpatialBot in Embodiment Tasks}
\begin{figure}[h]
	\begin{center}
    \includegraphics[width=1\linewidth]{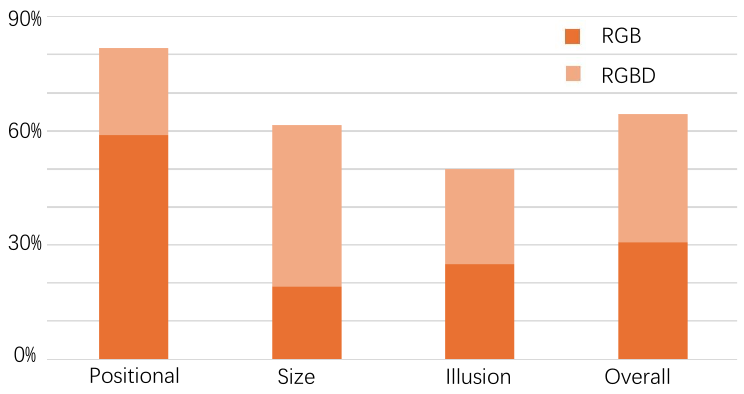}
	\end{center}
 	\caption{\textit{SpatialBot} success rate in pick-and-place of RGB and RGBD variants.}
	\label{fig:embodiment-success-rate}
\end{figure}

We finetune \textit{SpatialBot} on \textit{SpatialQA-E} to do manipulation tasks on real robots. It can be seen as a VLA model supporting multi-frame RGB or RGBD inputs. 
We use QWen-1.5-0.5B~\cite{qwen} as the base LLM and CLIP~\cite{clip} as the vision encoder. The pretrain dataset is Bunny-pretrain-LAION-2M~\cite{bunny}, and \textit{SpatialQA-E} is used in finetuning. Four frames in history are used to predict the end-effector delta position of the current frame. The model runs locally or connects through an ssh/sftp connection to run on RTX 4090 GPU. It is validated through experiments that \textit{SpatialBot} can do manipulation tasks with spatial instructions. Fig.~\ref{fig:embodiment-success-rate} shows the success rate of \textit{SpatialBot} RGB and RGBD variants. With depth information, \textit{SpatialBot} can pick and place more accurately.

\section{Monocular Depth Estimation by VLM}
We let VLMs to understand depth input so as to get accurate depth information. Readers may wonder that, is MDE for a single point or object difficult? To verify this, we ask \textit{SpatialBot} to predict the depth of points or objects. 
The results are reported in Table. 1 of main paper (Depth scores of \textit{SpatialBot}-RGB).
It seems that VLMs has not been fully prepared for MDE in a text-only output fashion. As reported in~\cite{mllmmde1,mllmmde2,mllmmde3,mllmmde4}, an extra decoder may be needed to generate accurate and pixel-level depth. We are not doing so since this new structure limits the generality of SpatialQA.

\section{SpatialBench Metadata}
In SpatialBench, GPT-4o generate 34 multiple choice questions regarding positional relationships, and we use the correct selection ratio as accuracy. 
GPT-4o generated 20 positive and negative question pairs on another 20 image. Only when the model answers both the positive and negative questions of a problem correctly is it considered correct.
On 20 image, human experts choose a category of object and annotate its quantity. We ask the model to count the objects.
Human experts think of positive and negative multiple choice questions regarding whether object A has reached or touched object B, and size comparisons on 20 images respectively. We first calculate the rate of correct choices from models. When it answers a pair of positive and negative questions correctly, we give it a bonus score.

\section{Data Generation in SpatialQA}
\begin{figure*}[h]
	\begin{center}
    \includegraphics[width=1\linewidth]{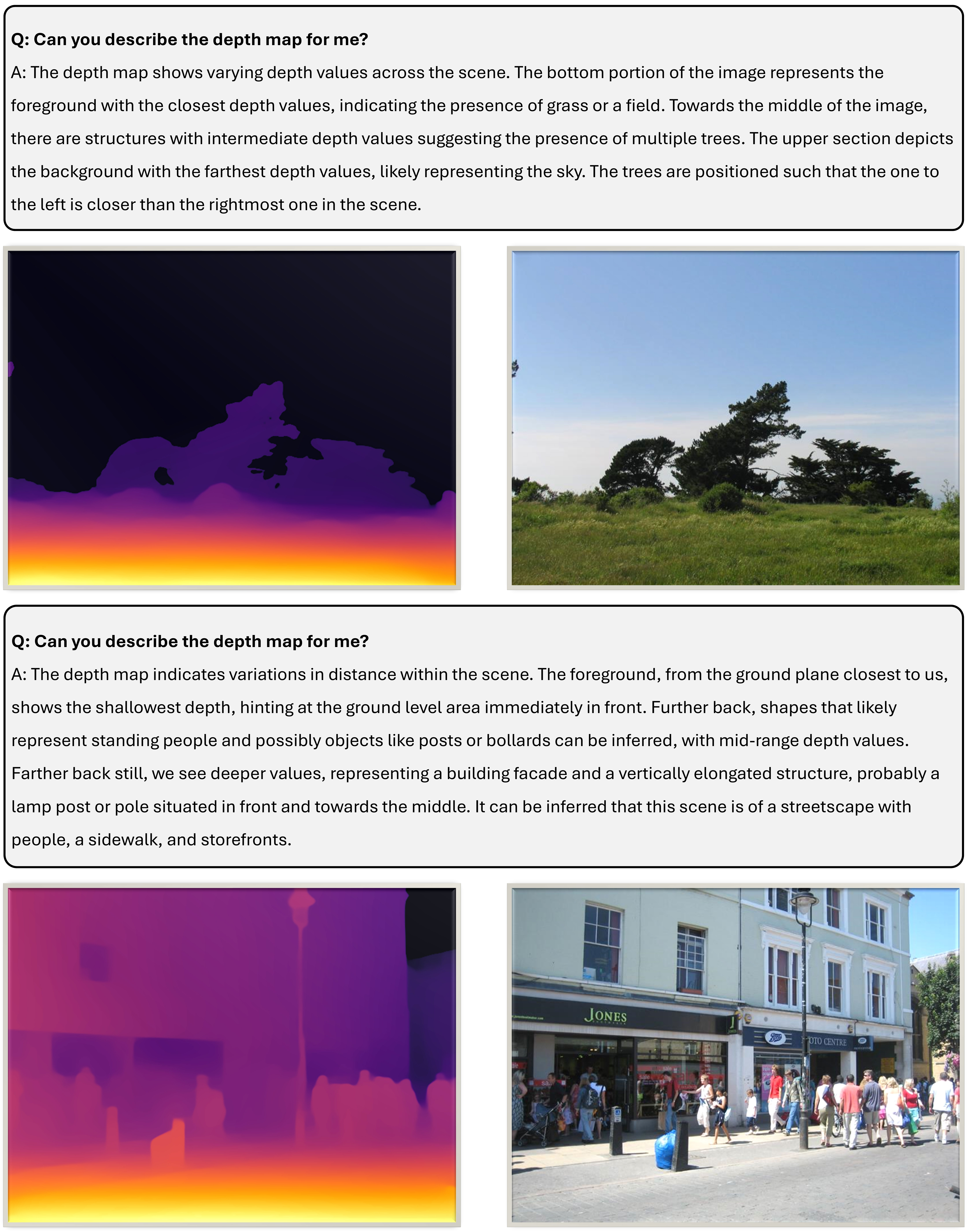}
	\end{center}
 	\caption{Sample data of depthmap understanding in \textit{SpatialQA}.}
	\label{fig:spatialqa_samples_depth}
\end{figure*}

\begin{figure*}[h]
	\begin{center}
    \includegraphics[width=1\linewidth]{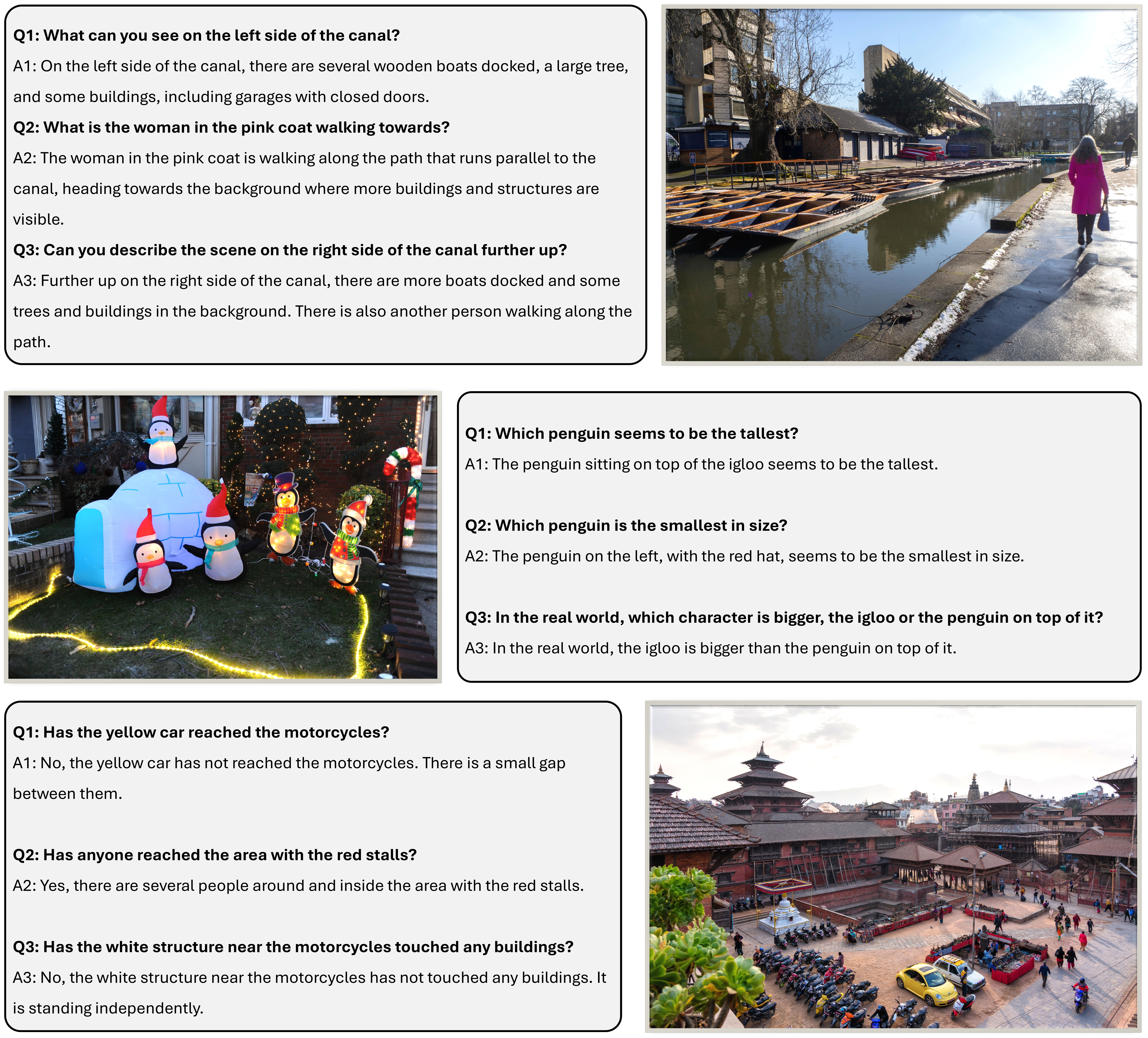}
	\end{center}
 	\caption{Sample data of spatial understanding in \textit{SpatialQA}.}
	\label{fig:spatialqa_samples_spatial}
\end{figure*}

\begin{figure*}[h]
	\begin{center}
    \includegraphics[width=1\linewidth]{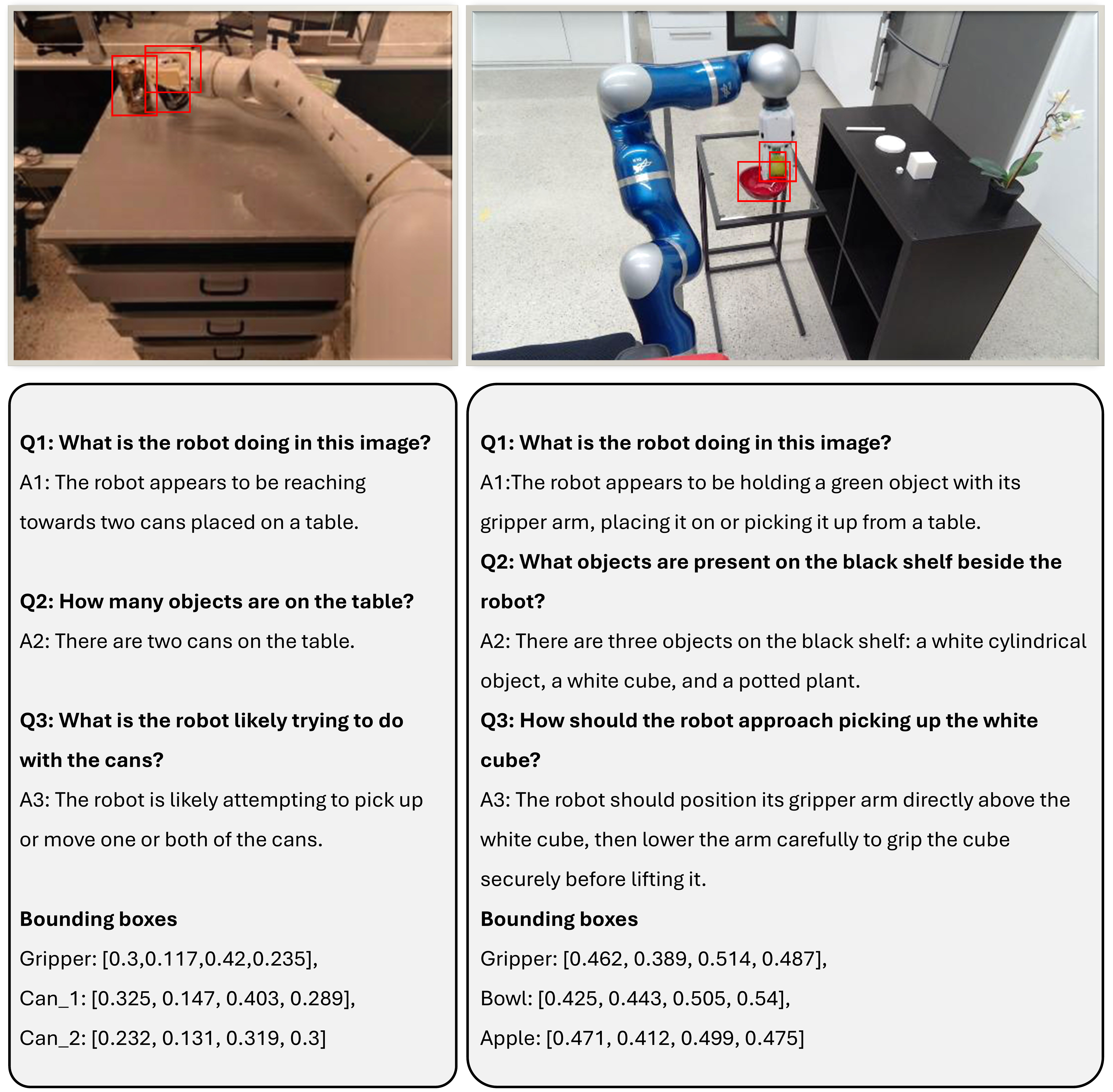}
	\end{center}
 	\caption{Sample data of robot scenes in \textit{SpatialQA}.}
	\label{fig:spatialqa_samples_robot}
\end{figure*}

We prompt GPT on about 50k images for depthmap understanding, spatial understanding and robot scene understanding in SpatialQA. The data source and data selection protocol are shown in Table.~\ref{tab:img_source}. GPT prompts we use for three seperate tasks are shown in Table.~\ref{tab:gpt_prompt}. Sample images and generated QAs are shown in Fig.~\ref{fig:spatialqa_samples_depth},~\ref{fig:spatialqa_samples_spatial},~\ref{fig:spatialqa_samples_robot}. 
Additionally, human experts annotated 3 bounding boxes per image in RTX, and are asked to annotate gripper if it is clearly visible in the image.

\section{Dataset, Model and Benchmark Usage}
The official repository of SpatialBot is \url{https://github.com/BAAI-DCAI/SpatialBot?tab=readme-ov-file}, where we provide metadata, codes, scripts, checkpoints, licenses and links to resources.

SpatialQA can be downloaded on Hugging Face: \url{https://huggingface.co/datasets/RussRobin/SpatialQA}.

SpatialQA-E can is available on Hugging Face: \url{https://huggingface.co/datasets/RussRobin/SpatialQA-E}. 

SpatialBench can be accessed through \url{https://huggingface.co/datasets/RussRobin/SpatialBench}.

Checkpoint for SpatialBot-3B, which is based on Phi-2~\cite{phi3} and SigLip~\cite{siglip}, can be downloaded at \url{https://huggingface.co/RussRobin/SpatialBot-3B}.

The pretrained checkpoints of SpatialBot can be accessed through Bunny~\cite{bunny} model zoo: \url{https://github.com/BAAI-DCAI/Bunny}.

\begin{table*}[ht]
      \caption{RGB image and depth sources in SpatialQA. Depthmap is either from sensors, as included in the original datasets, or MDE depth by ZoeDepth~\cite{zoe}.}
    \small
    \centering
    \begin{tabular}{p{3cm}|p{3cm}|p{2cm}|p{9cm}}
     \toprule
        \textbf{Data} & \textbf{Aim} & \textbf{Img. Num.} & \textbf{Image Selection Protocol and Depthmap Source} \\ \midrule
        Bunny695k~\cite{bunny} & General MLLM abilities & 695k & - \\ \midrule
        VG, COCO & depthmap understanding & 20k & Random selection. GPT is prompted to first infer from the depth color map and then verify its inference against the RGB data to ensure its correctness. The inference of GPT is stored in QA format. MDE.\\ \midrule
        KITTI~\cite{kitti} & spatial understanding & 1.75k & Randomly select images from each sequence of the KITTI dataset. Since some scenes are captured when the car is stationary or temporarily stopped (e.g., at a traffic light or due to a stopped vehicle ahead), there are many repetitive images. We then Manually filter out those with a particularly high repetition rate. Since depth information in KITTI only include sparse points, and ZoeDepth has been finetuned on KITTI, we use MDE. \\ \midrule
        NYU Depth v2~\cite{nyudepthv2} & spatial understanding & 1.5k & All images and sensor depth images are adopted. \\ \midrule
        RT-X~\cite{rtx} & robot scene understanding & 7.5k & We annotate 3 bounding boxes per image, and annotate the gripper if it is visible. We also prompt GPT to generate general QAs. We use depthmap from subsets if available. Otherwise, we use MDE to estimate depth.\\ \midrule
        SA-1B~\cite{sam} & spatial understanding & 15k & We randomly select real-world images from SA-1B and prompt GPT-4o to generate conversations regarding spatial relationships. MDE.\\ \midrule
        2D-3D-S~\cite{2d3ds} & spatial understanding & 2.9k & We randomly select images from 2d3ds and manually exclude images with no more than 3 objects in them. Sensor depth. \\ \bottomrule
        \end{tabular}
    \label{tab:img_source}
\end{table*}

\begin{table*}[ht]
    \caption{GPT prompts used in SpatialQA.}
    \small
    \centering
    \begin{tabular}{p{4cm}|p{13cm}}
     \toprule
        \textbf{Aim} & \textbf{Prompts for GPT} \\ \midrule
        depthmap understanding & Design a conversion between you and a human talking about the depth map. The human asks you to describe the depth map. You should focus on depth value predictions. The colors just represent depth values. Do not directly mention colors on the image in your response, instead, mention the depth distribution they stand for. Looking at the depth map, you should also infer what may be in the image. If something really exists in the rgb image, and can be inferred from the depth map, you can mention they in your response. If possible, pay attention to spatial relationships. When referring to spatial relationships, such as left and right, you should use the real-world left and right, rather than those in the image coordinate system. \\ \midrule
        
        spatial understanding & Design a conversation, consisting of no more than 3 Question-Answer pairs, between you and a person asking about this image. The content within the conversation should be logically connected. You should think of what are spatial relationships of objects in the image. Then generate the conversation according regarding the spatial relationships. Spatial relationships can be about, but not limited to these categories: positional (left/right, below/above, behind/front), distance (further/closer to the camera, further/closer to something), size(big/small, tall/short, wide/thin), reach (has A touched/reached B physically). When describing spatial relationships, always use the real-world orientation as if you are standing in the real scene. e.g. when using right side of object, it should talk about what is on the right side of object in the real world, not on the right side of image'. Only describe the things that you are sure about. \\ \midrule

        robot scene understanding & Design a conversation, consisting of no more than 3 Question-Answer pairs, between you and a person asking about this image. The content within the conversation should be logically connected. You should first think of robot task: what may the robot want to do with the objects. And then generate the conversation according to robot task. The conversations can include what robot are doing, how should the robot finish robot task, object count, object position, positional relationships, object appearance, etc. Only describe the things that you are sure about. Please note that you are talking to a person about the image and robot. You are not the robot, and you are not talking to the robot. \\ \bottomrule
        \end{tabular}
    
    \label{tab:gpt_prompt}
\end{table*}

\section{Conclusion}
\label{sec:conclusion}
We propose \textit{SpatialBot}, a family of state-of-the-art VLMs, for effective depth understanding and thus precise robot manipulating in embodied AI by training on our constructed \textit{SpatialQA} and \textit{SpatialQA-E} datasets. 
\textit{SpatialBot} can understand depth inputs and use depth information to do spatial understanding and reasoning tasks in Visual QA and Embodiment.
\textit{SpatialBench} is also designed to evaluate the model performance of spatial knowledge in multiple aspects.
Experimental results on our benchmark, general VLM benchmarks, and robot manipulation deployment verify the effectiveness and superiority of \textit{SpatialBot} comparing to competitors.

\section*{Acknowledgement}
This work was supported by National Natural Science Foundation of China under No. 62306046 and the National Youth Talent Support Program under No. 8200800081.

\clearpage

{
    \small
    \bibliographystyle{ieeenat_fullname}
    \bibliography{main}
}

\end{document}